\documentclass[11pt]{article}

\usepackage[margin=1in]{geometry}

\usepackage[utf8]{inputenc}
\usepackage[T1,T2A]{fontenc}
\usepackage[main=english,russian]{babel}

\usepackage{amsmath,amssymb}
\usepackage{graphicx}
\usepackage{float}
\usepackage{tabularx}
\usepackage{array}
\usepackage{booktabs}
\usepackage{longtable}
\usepackage{url}
\usepackage[hidelinks]{hyperref}

\newcolumntype{Y}{>{\raggedright\arraybackslash}X}

\title{Markov reads Puškin, again: A statistical journey into the poetic world of Evgenij Onegin}
\author{Angelo Maria Sabatini\\
The BioRobotics Institute\\
Scuola Superiore Sant'Anna\\
Pisa, Italy\\
\texttt{angelo.sabatini@santannapisa.it}}
\date{}

\begin{document}
\selectlanguage{english}
\maketitle

\section*{Abstract}
This study applies symbolic time series analysis and Markov modeling to explore the phonological structure of \textit{Evgenij Onegin}---as captured through a graphemic vowel/consonant (V/C) encoding---and one contemporary Italian translation. Using a binary encoding inspired by Markov’s original scheme, we construct minimalist probabilistic models that capture both local V/C dependencies and large--scale sequential patterns. A compact four-state Markov chain is shown to be descriptively accurate and generative, reproducing key features of the original sequences such as autocorrelation and memory depth. All findings are exploratory in nature and aim to highlight structural regularities while suggesting hypotheses about underlying narrative dynamics.

The analysis reveals a marked asymmetry between the Russian and Italian texts: the original exhibits a gradual decline in memory depth, whereas the translation maintains a more uniform profile. To further investigate this divergence, we introduce phonological probes~\textemdash\ short symbolic patterns that link surface structure to narrative-relevant cues. Tracked across the unfolding text, these probes reveal subtle connections between graphemic form and thematic development, particularly in the Russian original.

By revisiting Markov’s original proposal of applying symbolic analysis to a literary text and pairing it with contemporary tools from computational statistics and data science, this study shows that even minimalist Markov models can support exploratory analysis of complex poetic material. When complemented by a coarse layer of linguistic annotation, such models provide a general framework for comparative poetics and demonstrate that stylized structural patterns remain accessible through simple representations grounded in linguistic form.

\section*{Introduction}
Few ideas in modern science have enjoyed the reach and longevity of the Markov chain. Originally introduced by the Russian mathematician Andrej A. Markov, it describes a stochastic process in which a system transitions between states according to probabilistic rules that depend only on the current state. The set of states may be directly observable by an external observer, or hidden, with observable outputs generated through conditional probability distributions; this is the case of a Hidden Markov Model (HMM)~\textemdash\ a doubly stochastic process in which both state transitions and output emissions are governed by probabilistic laws.

Although the simplest Markov models rely on the assumption that only the present state influences the next (a property often referred to as memorylessness), this condition can be generalized to allow finite-order dependencies on the recent past. In such cases, the chain is said to be of order $m$, meaning that the probability of transitioning to the next state depends on the $m$ most recent states (with $m$ finite). 

Markov chains underpin applications ranging from speech recognition \cite{rabiner1989} and web search \cite{ilprints422}, to computational biology \cite{krogh1994}, finance \cite{siu2007}, statistical physics, where they are widely used to study collective behavior in systems of interacting particles \cite{propp1996exact}, and even biomedical contexts \cite{newton2012, mannini2012}.

However, it is not widely known that the origins of Markov chains lie far from the domains in which they later gained prominence. Their development stemmed from the mathematical interests of Markov himself, who sought to demonstrate that the weak law of large numbers~\textemdash\ traditionally formulated for sequences of independent random variables~\textemdash\ could be extended to certain sequences exhibiting statistical dependence \cite{markov1906}.

To advance his argument, Markov turned to an empirical domain as unexpected as it was rich: literature. By encoding the first 76 stanzas of \textit{Evgenij Onegin}~\textemdash\ the canonical novel in verse by Aleksandr S. Puškin~\textemdash\ as a binary sequence of 20{,}000 characters representing vowels and consonants, he constructed a two-state process with explicit dependence between elements. In doing so, he effectively introduced what we would now describe as a two-state Markov chain; in particular, he succeeded in showing that the law of large numbers could hold even in the presence of sequential graphemic dependencies, namely statistical correlations between adjacent vowels and consonants \cite{markov1913}.

His 1913 paper marks the birth of the Markov chain as a mathematical object, although the term itself would enter the lexicon only years later (see \cite{seneta1996} for a comprehensive historical account). In making this move, Markov not only challenged prevailing assumptions about independence, but also inaugurated what has been described as the \textit{mathematization of writing}~\textemdash\ a gesture that brought statistical reasoning into direct contact with linguistic form \cite{link2006}.

It is now well understood that language exhibits a dual nature: on the one hand, structural constraints imposed by the vocal tract, cognitive architecture, and communicative norms; on the other, stochastic variability shaped by social, historical, and individual contingencies. Whether or not Markov explicitly theorized this duality, his method anticipated it with clarity; it yielded, among other things, statistical support for the empirical fact that in natural language, the probability of a vowel following a consonant (and vice versa) is elevated due to the constraints of human articulation.

Markov himself extended his investigations by encoding a short novel by Sergej T. Aksakov as a time series of alternating vowels and consonants comprising up to 100{,}000 characters, in order to validate the observed dependencies within a more homogeneous sample of prose. However, the specific application of Markov chains to literary analysis did not gain significant traction in the immediate aftermath \cite{basharin2004life}. During Markov’s lifetime, the only notable continuation of textual analysis was an attempt to construct linguistic spectra based on word frequency statistics~\textemdash\ a pioneering effort to address questions of authorship attribution in a quantitative framework \cite{morozov1915}. 

It was in the mid-20th century that Claude Shannon, working in a radically different context, drew upon Markovian ideas to formulate a theory of communication \cite{shannon1948}. He modeled text as a stochastic source from which statistical properties such as entropy, redundancy, and channel capacity could be mathematically derived. Yet Shannon’s goal was not to analyze literature per se; rather he was interested in investigating synthetic generation \cite{link2006b}. He famously constructed pseudo-English sentences based on conditional probabilities of increasing order; they were not intended as literary artifacts, but as demonstrations of how statistical structure alone can create the illusion of meaning. These experiments echoed Markov’s original gesture, while shifting the emphasis from analysis to generation \cite{hayes2013}.

Stylometry~\textemdash\ the quantitative analysis of literary style~\textemdash\ has largely developed along an independent trajectory, focusing on lexical, syntactic, and statistical features for tasks such as authorship attribution and genre classification \cite{stamatatos2009survey}. An interesting exception is the approach proposed in \cite{khmelev2001}, where a technique for authorship attribution was based on a simple Markov chain of letters, that is, only letter bigrams, anticipating later explorations of sublexical structure. In recent years a growing body of research has continued to explore the phonological and rhythmic dimensions of literary texts, often drawing on techniques from signal processing, information theory, and computational linguistics \cite{sharma2024deep}. Notable examples include a study comparing features of Puškin’s poetic style with those of his contemporaries using machine learning methods \cite{barakhnin2022}, and an investigation of information entropy in poetic texts, again carried out through comparative analysis of Puškin and other poets of the Russian Golden Age \cite{kozhemyakina2023}.

Apart from \cite{khmelev2001} and a few other developments \cite{petruszewycz1983}, the direct line of influence stemming from \cite{markov1913} did not lead to a sustained tradition in the analysis of literary texts. What remained, however, was a deeper and more general insight: a text can be modeled as a symbolic time series, and its mathematical structure can be analyzed, regardless of whether the assumptions of Markovian memory strictly hold \cite{link2006b}. As observed in \cite{petruszewycz1983}, Markov chains were often criticized and rarely adopted in linguistic studies. Yet it is also suggested there that such methods might be purposefully revisited, for instance, in relation to alternative approaches to text segmentation or for the comparative analysis of the same text across different languages.

Within this broader landscape, the present study revisits Markov’s original gesture not as a historical curiosity, but as a methodological opportunity: to examine the structural dynamics of a literary masterpiece using minimal probabilistic models and a symbolic encoding that may help foreground aspects of its underlying vowel--consonant (V/C) scaffolding. By reanalyzing \textit{Evgenij Onegin} in its entirety, along with one of its Italian translations, through the lens of modern computational statistics and data science, we examine how V/C structures may reflect stylistic tendencies. We further show how Markov’s foundational insight can be revisited within a contemporary analytical framework, more than a century after its original formulation.

\subsection*{\textit{Evgenij Onegin}}
More than a literary masterpiece, \textit{Evgenij Onegin} represents the founding act of the modern Russian literary language. It has shaped generations of Russian poets and writers, and continues to stand as a cultural and linguistic landmark. Composed between 1823 and 1831, the poem is divided into eight chapters, each comprising between 46 and 60 stanzas, for a total of 421.

The minimalist narrative of \textit{Evgenij Onegin} traces the emotional trajectory of its eponymous protagonist, a young aristocrat who inherits an estate and moves to the countryside, as described in Part 1. There, he befriends the poet Vladimir Lenskij and meets Tat’jana Larina, the other main character of the poem~\textemdash\ a shy and introspective young woman who soon falls in love with him (Parts 2-3). Evgenij firmly rejects her in Part 4. Later, in a thoughtless moment, he flirts with Olga (Tat’jana’s sister and Lenskij’s fiancée), setting in motion the events of Part 5 and culminating in a fatal duel in Part 6, when Lenskij succumbs to death.

After years of forced absence and seemingly aimless wandering, Evgenij unexpectedly encounters Tat’jana again~\textemdash\ now transformed (Part 7). Her astonishing evolution~\textemdash\ from a quiet, imaginative girl of the rural nobility to a prominent figure in St. Petersburg salons~\textemdash\ culminates in a marriage not born of passion, but of duty. A crucial moment in this transformation is her solitary visit to Evgenij’s home during his absence, where she engages intimately with his inner world through his books and belongings (also in Part 7). Deeply struck by her poise and presence, Evgenij falls in love with her and declares his feelings in Part 8. Although her love for him endures, she turns him away with composure, leaving him suspended in a narrative that Puškin deliberately chooses not to complete. The emotional arc of the poem unfolds within what can be described as a world structured by social conventions, yet constantly destabilized by the unpredictability of chance and inner transformation \cite{lotman1976}.

A distinctive feature of the poem lies in its formal invention. Puškin devised a unique stanzaic form, now known as the Onegin stanza, consisting of 14 lines in iambic tetrameter with a fixed rhyme scheme (AbAbCCddEffEgg). This structure yields lines of nearly uniform syllabic length. Yet within this strict metrical constraint, Puškin achieves extraordinary expressive range. The poem abounds in subtle shifts of tone, register, and diction, often within a single stanza, and exhibits masterful rhythmic control through devices such as pyrrhic substitutions, which momentarily disrupt metrical expectations while preserving the underlying iambic pulse \cite{pushkin2021}.

The Puškin’s linguistic versatility is remarkable: the Russian original displays a high heterogeneity of register, blending colloquial idioms, poetic archaisms, neologisms, and expressions from liturgical, military, bureaucratic, and popular jargons, alongside foreign borrowings from English, French, Italian, and German. This linguistic mélange reinforces the poem’s tonal and social dynamism, while enriching its stylistic texture and rhythmic complexity.

The expressive variability of the poem arises in part from the subtle interplay between meter and rhythm. While the metrical scheme remains constant~\textemdash\ namely, four iambic feet per line~\textemdash\ the placement and type of stress-bearing words vary. As Russian verse theory suggests, and as Tomaševskij famously argued, rhythm emerges not merely from the metrical grid but from the concrete realization of stress patterns in each line: meter is an abstract scheme, and rhythm is the actual sound form~\textemdash\ the concrete arrangement of stresses in each individual line \cite{pilshchikov2019}. These so-called rhythmic elements, typically the final stressed word or phrase in each line, can be classified as masculine (final-syllable stress) or feminine (penultimate-syllable stress), and their alternation contributes to the poem’s deeper prosodic texture.

To accompany the analysis of the Russian original, we selected the recent Italian translation by Giuseppe Ghini \cite{pushkin2021}. While the language of the translation is not central to our approach, Ghini’s version is notable for its formal fidelity and expressive nuance. It replicates Puškin’s stanzaic form while deliberately discarding the original rhyme scheme in order to preserve the rhythmic cadence and tonal complexity of the Russian text. His use of unrhymed nine-syllable lines (novenari sciolti) maintains a metrically coherent structure, allowing for stylistic agility and prosodic balance.

Our present study is guided by the hypothesis that stylistic and structural signals~\textemdash\ typically sought in lexical or syntactic patterns~\textemdash\ may also emerge from the basic alternation of vowels and consonants. By modeling the texts as symbolic time series (i.e., sequences of categorical observations~\textemdash\ vowels and consonants~\textemdash\ indexed by position in the text), we investigate the extent to which graphemic structure may reflect deeper aspects of literary design. Markov’s original analysis adopted a simplified binary encoding of vowels and consonants, disregarding features such as palatalization signs and word boundaries. Though later criticized from a philological standpoint, this abstraction proved foundational in modeling symbolic linguistic structure, and it continues to offer valuable insight \cite{petruszewycz1983}~\textemdash\ an approach we adopt and extend in the present study.

The rest of the paper is organized as follows. The next section describes the dataset and preprocessing steps, including text segmentation, binary encoding, and structural alignment between the Russian original and its Italian translation. We then introduce the Markov modeling framework and describe the metrics that are employed to validate the model and to assess the evolution of graphemic dependencies across the text. The results section presents our empirical findings, including graphemic trends over large text segments, trigram analyses, and a focused investigation of recurring vowel and consonant patterns in context, referred to here as phonological probing. We conclude with a discussion of the methodological implications and potential extensions to cross-linguistic and stylometric research.

\section*{Materials and methods}
\subsection*{Corpus composition}
We analyzed a bilingual corpus of \textit{Evgenij Onegin}, retaining only the eight canonical chapters of the main poetic body and excluding the dedication and appendices. The Russian text, sourced from Litra (https://www.litra.ru), follows post-1918 orthographic conventions. Although the provenance of this digital version is undocumented, it was systematically checked against the printed edition used by Ghini~\textemdash\ which itself follows the \textit{Polnoe Sobranie Sočinenij} (16 vols., Nauka, Leningrad, 1937-49)~\textemdash\ through spot-checking multiple stanzas across all chapters. While a full collation was not performed, the examined passages exhibit a high degree of correspondence, with only minor typographic discrepancies (e.g., ASCII double quotes replacing guillemets for marking direct speech and em-dashes incorrectly rendered as en-dashes), ensuring sufficient consistency for structural alignment. This internal consistency of the Russian corpus is further supported by the close agreement between the vowel-consonant unigram, bigram, and trigram counts obtained here on the initial stanzas and those reported in Markov’s original analysis (see Section “Vowel-consonant encoding and Markov modeling”). The Italian translation was scanned and processed using \texttt{ocrmypdf}, an open-source tool that applies Optical Character Recognition (OCR) to PDF documents and embeds the recognized text as an invisible layer. The procedure was performed under macOS and subsequently verified manually against the printed edition. Corrections primarily addressed minor typographic or OCR-related issues. Given the deliberately coarse nature of the vowel-consonant encoding and the sizes of the textual blocks involved in our proposed framework, isolated character-level OCR errors are not expected to propagate systematically or to bias blockwise vowel-consonant counts or local co-occurrence patterns.

\subsection*{Text alignment and encoding}
All preprocessing and analyses were performed in R (version 4.4.1). 

We developed a parser to perform stanza-level alignment of the texts, using the Russian structure as the primary reference and matching the corresponding Italian segments accordingly. This alignment preserves the poem’s formal organization and enables per-stanza comparisons across versions. Internal line segmentation was retained to allow for finer-grained analyses. Chapter-opening epigraphs were annotated separately and excluded from quantitative modeling unless explicitly stated. Structural anomalies~\textemdash\ namely, empty lines filled with dots (sometimes replacing entire stanzas) or the fusion of multiple stanzas into a single unit~\textemdash\ were likewise preserved and annotated, as they reflect deliberate poetic strategies, including, for instance, attempts to obscure politically sensitive content.

To replicate and extend Markov’s original analysis, text was encoded as a binary vowel-consonant sequence, following his original exclusion of punctuation, whitespace, and the hard (ъ) and soft (ь) signs of the Cyrillic alphabet. The encoding was adapted to post-1918 orthography and extended to include Latin-script foreign expressions present in the Russian text. A detailed classification scheme, including edge cases, is provided in S1 Appendix. In addition, structural features such as per-line character and word counts were computed to support blockwise segmentation and stylometric analyses.

\subsection*{Vowel-consonant encoding and Markov modeling}
Each encoded text was modeled as a symbolic time series. Two models were considered, in line with standard formulations in stochastic modeling and symbolic dynamics \cite{markov1913, petruszewycz1983, dobrow2016}. The first is a classical two-state model, where each character is either V (vowel) or C (consonant), and transitions are governed by first-order probabilities:

\begin{equation*}
\begin{aligned}
  p_0&=P(V\mid C);\;\;q_0=P(C\mid C)\\
  p_1&=P(V\mid V);\;\;q_1=P(C\mid V)
\end{aligned}
\end{equation*}

The second model captures second-order dependencies by expanding the state space to four bigram-based states: VV, VC, CV, and CC. Although this formulation remains formally first-order~\textemdash\ since transitions depend only on the current state~\textemdash\ it embeds a two-symbol memory by representing overlapping pairs of V/C symbols as states. The transition probabilities then encode the likelihood of observing the next symbol given the previous two (see Table~\ref{tab:tab-tab1}). The probabilities of transitioning to a consonant, given each bigram state, are $q_{ij} = 1 - p_{ij}$ with $i,j \in \{0,1\}$. Each transition shifts the bigram forward, dropping the older symbol and appending the newly generated one.

\begin{table}[!ht]
\centering
\caption{{\bf Transition probabilities in the four-state Markov model.}
Each state corresponds to a bigram of the previous two symbols; transition probabilities determine the likelihood of observing a vowel.}
\smallskip
\label{tab:tab-tab1}
\footnotesize
\renewcommand{\arraystretch}{1.1}
\begin{tabular}{lll}
\toprule
\textbf{Current state} & \textbf{Previous two symbols} & \textbf{Probability next symbol} \\
\midrule
$VV$ & $(X_{t-2}=V, X_{t-1}=V)$ & $p_{11} = P(V \mid VV)$ \\
$VC$ & $(X_{t-2}=V, X_{t-1}=C)$ & $p_{10} = P(V \mid VC)$ \\
$CV$ & $(X_{t-2}=C, X_{t-1}=V)$ & $p_{01} = P(V \mid CV)$ \\
$CC$ & $(X_{t-2}=C, X_{t-1}=C)$ & $p_{00} = P(V \mid CC)$ \\
\bottomrule
\end{tabular}
\end{table}

\begin{figure}[!h]
\centering
\includegraphics[width=0.9\linewidth]{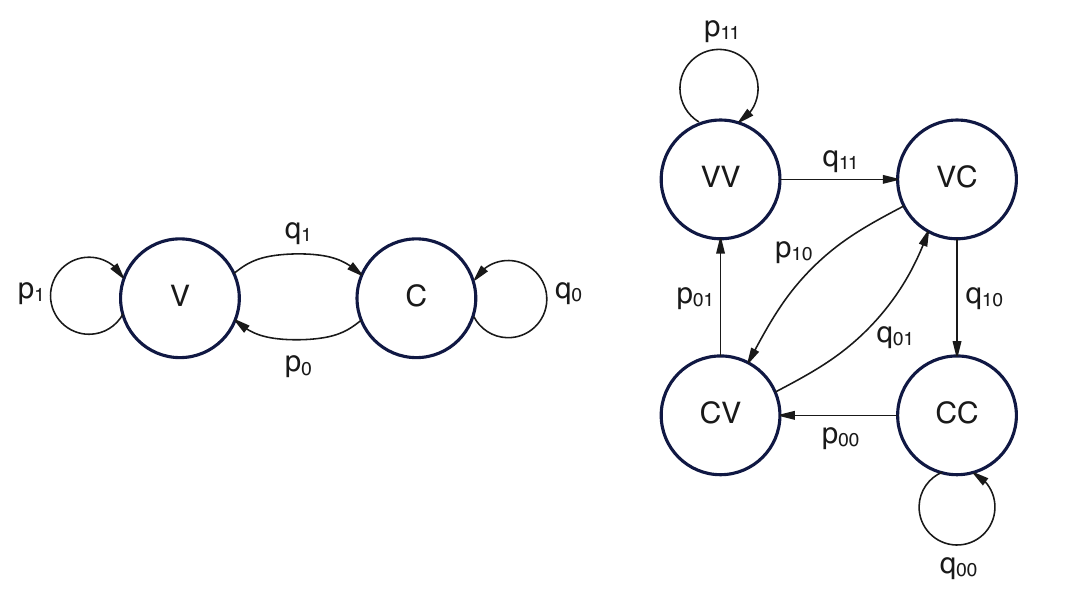}
\caption{{\bf Graph representation of the Markov models for vowel-consonant alternation.}
The two-state model (left) tracks transitions between individual symbols (V, C), while the four-state model (right) captures dependencies via bigram-based states (VV, VC, CV, CC).}
\label{fig:fig-1}
\end{figure}

While retaining the formal simplicity of a first-order chain, the four-state model allows for more nuanced modeling of local dependencies by encoding a deeper memory. This enriched representation underlies both the simulation procedures and the computation of so-called dispersion coefficients (see below). The structure of both Markov chain types is summarized in Fig~\ref{fig:fig-1}.

The V/C encoding is defined at the graphemic level and can be interpreted as a coarse proxy for phonological structure. To quantify the statistical consequences of symbol-level dependencies in the binary V/C sequence, we introduced a correction factor $\text{CF}$ (also known as the dispersion coefficient in Markov’s original terminology), which measures the inflation or reduction in variance due to sequential dependence.

\begin{equation*}
\sigma^2_{\text{independent}}=\frac{p(1-p)}{n}\quad\Rightarrow\quad\sigma^2_{\text{dependent}}=\text{CF}\cdot \sigma^2_{\text{independent}}
\end{equation*}

Two such correction factors can be defined:

- simple correction factor, $\text{CF}_{\text{simple}}$, derived from the lag-1 autocorrelation $d$ of the binary sequence:

\begin{equation}
\text{CF}_{\text{simple}}=\frac{1+d}{1-d} 
\label{eq:CF_simple}
\end{equation}

with

\begin{equation}
d=p_1-p_0
\end{equation}

- complex correction factor, $\text{CF}_{\text{complex}}$, derived from the full structure of the four-state model:

\begin{equation}
\text{CF}_{\text{complex}}=\dfrac{1}{2}\left[\dfrac{1+\eta}{1-\eta}+\dfrac{1+\nu}{1-\nu}\right] \text{CF}_{\text{simple}}+\dfrac{(q-p)(\nu -\eta)}{(1-\eta)(1-\nu)}
\end{equation}

where $p = P(V)$ and $q = P(C)$ denote the stationary probabilities of vowels and consonants, respectively, with

\begin{equation}
\begin{aligned}
\eta&=\frac{p_{11}-p_1}{q_1}\\
\nu&=\frac{q_{00}-q_0}{p_0}
\end{aligned}
\end{equation}

The quantities $\eta$ and $\nu$ were defined to measure how much the conditional probability of a symbol (V or C) changes when extending the context from one to two preceding symbols. The normalization accounts for the maximum possible change, namely $q_1$ for vowels and $p_0$ for consonants.

In the present context, both $\eta$ and $\nu$ are negative, reflecting a systematic reduction in predictive strength when moving from bigrams to trigrams. Applying the Cyrillic V/C encoding to the first 76 stanzas of the modernized Russian text, we obtained $\text{CF}_{\text{simple}} = 0.303$ and $\text{CF}_{\text{complex}} = 0.199$ ($\eta = -0.021$, $\nu = -0.297$). These values closely match those originally reported by Markov in \cite{markov1913} (0.303 and 0.195; –0.027 and –0.309, respectively), see S1 Appendix for details. In the remainder of the paper, unless otherwise noted, $\text{CF}$ refers to the complex correction factor $\text{CF}_{\text{complex}}$. To facilitate interpretation and comparative analysis, we also define the memory depth ($\text{MD}$):

\begin{equation}
\text{MD} = 1 - \text{CF}
\label{eq:MD}
\end{equation}

$\text{MD}$ ranges from 0 (perfect dependence) to 1 (perfect independence), with higher values indicating greater structural randomness. This index serves as the primary outcome in our block-level analysis to follow. MD should be understood as a descriptive index summarizing the extent to which higher-order conditioning improves local predictability in a finite-order Markov representation \cite{gagniuc2017}. As such, MD is conceptually related to notions of effective memory length, rather than to asymptotic properties such as entropy rate or long-range mixing. It provides a convenient measure of how strongly local context contributes to short-range structure in symbolic sequences.

\subsection*{Statistical modeling and computational procedures}
Building on the encoded V/C sequences described above, all subsequent analyses were conducted using the four-state Markov model (VV, VC, CV, CC), which embeds second-order dependencies while retaining a first-order formulation. This representation supports the computation of the complex correction factor $\text{CF}$ and the derived memory depth index $\text{MD}$. Frequencies of bigrams and selected trigrams were also extracted to support model validation and narrative mapping.

To account for temporal dependence in binary sequences, we adopted the moving block bootstrap (MBB), a nonparametric resampling technique that preserves short-range correlations by drawing subblocks of fixed length with replacement from the observed data and concatenating them into new sequences \cite{kunsch1989, lahiri2003}. In our implementation, each original sequence was partitioned into consecutive, non-overlapping blocks of fixed length. Bootstrap replicates were generated independently for each block using subblocks of fixed size, which were randomly sampled with replacement from within the same block and concatenated in sequential order to preserve local temporal structure. A total of 1{,}000 replicates of the original sequence were thus produced and used in various stages of the analysis. Unless otherwise noted, all reported confidence intervals are derived from these bootstrap approximations.

The selected block and subblock lengths (10{,}000 and 250 characters, respectively) were empirically calibrated in pilot simulations to control the relative error in estimating $\text{CF}$. In addition, a minimal sensitivity analysis specifically assessing the robustness of block length selection is provided in S2 Appendix (Table~A1). Pilot simulations indicated that the chosen MBB configuration kept the relative estimation error within approximately 10\%, and even lower for $\text{MD}=1-\text{CF}$, thereby supporting reliable inference at the block level and enabling robust model comparison.

A surrogate-data procedure was also implemented \cite{theiler1992}. The original sequence was partitioned into subblocks whose length matched the subblock length used in the MBB; these subblocks were then randomly reshuffled across the entire sequence. This preserves local V/C structure while disrupting global sequential order. The resulting surrogates were processed through the same blockwise modeling pipeline and served as a null baseline to test the genuineness of observed trends in, e.g., $\text{MD}$, across the text.

To assess autocorrelation at multiple lags, we computed the autocorrelation function (ACF) for each block and applied the Ljung-Box test \cite{ljung1978} to evaluate the null hypothesis of independence. In addition, bootstrap confidence intervals for ACF values at lags 1 through 5 were derived using the MBB replicates generated for a single block. These estimates were then compared to the critical values obtained from the Ljung-Box test. Model adequacy was assessed by comparing empirical and simulated vowel-consonant trigram distributions. Discrepancy was quantified as the sum of absolute differences between empirical and simulated trigram relative frequencies. Median discrepancy values and bootstrap intervals were used to summarize variability across simulations.
 
We next examined how $\text{MD}$ co-varied with other linguistic properties across the text. This yielded a longitudinal profile of the corpus based on block-level measurements. In particular, we evaluated the association between $\text{MD}$ and relevant trigram frequencies using partial Spearman correlations.

To further assess the predictive contribution of language and position in the sequence, we fit a linear model including an interaction term between block index (position along the text) and source (Russian or Italian). Coefficients were estimated using ordinary least squares, and confidence intervals were obtained by bootstrapping the residuals within each block using the MBB replicates. As a diagnostic control, we applied a surrogate-data test to the Russian sequence: 250-character subblocks were reshuffled across the sequence to disrupt global order while preserving local structure. These surrogates were then processed through the same modeling pipeline, enabling direct comparison with the original data and testing whether the observed trends could be explained by short-range clustering alone.

Building on the longitudinal profiling results, we investigated higher-order phonotactic structure by extracting all occurrences of the trigrams VVV, VVC, CCC, and CCV in both corpora. Their distribution across blocks was examined using Spearman correlations to identify systematic trends.

We defined a phonological probe as a local pattern in the V/C sequence whose blockwise frequency profile may reflect underlying constraints in phonotactic structure. Trigrams exhibiting statistically consistent trends across blocks were selected as candidate probes and further examined through contextual retrieval.

Statistical modeling in this study was conducted within an exploratory observational framework, with inferential tools used for descriptive and diagnostic purposes rather than confirmatory hypothesis testing. Their role is to characterize associations, trends, and departures from simple null structures, and to highlight regions of potential stylistic or narrative relevance, rather than to establish causal relationships. The modeling approach is generative and descriptive, aimed at hypothesis generation rather than formal testing. This usage aligns with established recommendations for the application of inferential methods in observational and hypothesis-generating studies, where p-values and confidence intervals are interpreted as measures of compatibility with the data rather than as decision thresholds \cite{wasserstein2019}.

All analyses were performed in R, combining \texttt{tidyverse}–\texttt{tidytext} tools for structured data manipulation and tokenization, together with the \texttt{udpipe} library \cite{udpipe} for morphological annotation. Lemmatization and syntactic parsing were used to identify recurring lexical patterns associated with phonological probes (e.g., word groups containing the consonantal trigram "вст") and to assess their thematic consistency across the text. Scripts and summary data necessary for replication are available in the supplementary repository.

\section*{Results}
We began by analyzing basic structural features of the two texts, focusing on per-line counts of characters and words. The Russian text contains an average of $20.37 \pm 2.43$ characters per line ($4.28 \pm 1.06$ words), while the Italian translation averages $23.44 \pm 2.42$ characters ($5.12 \pm 1.15$ words). All these quantities follow approximately Gaussian distributions, as confirmed by visual inspection and standard normality tests (not shown).

\begin{figure}[!h]
\centering
\includegraphics[width=0.8\linewidth]{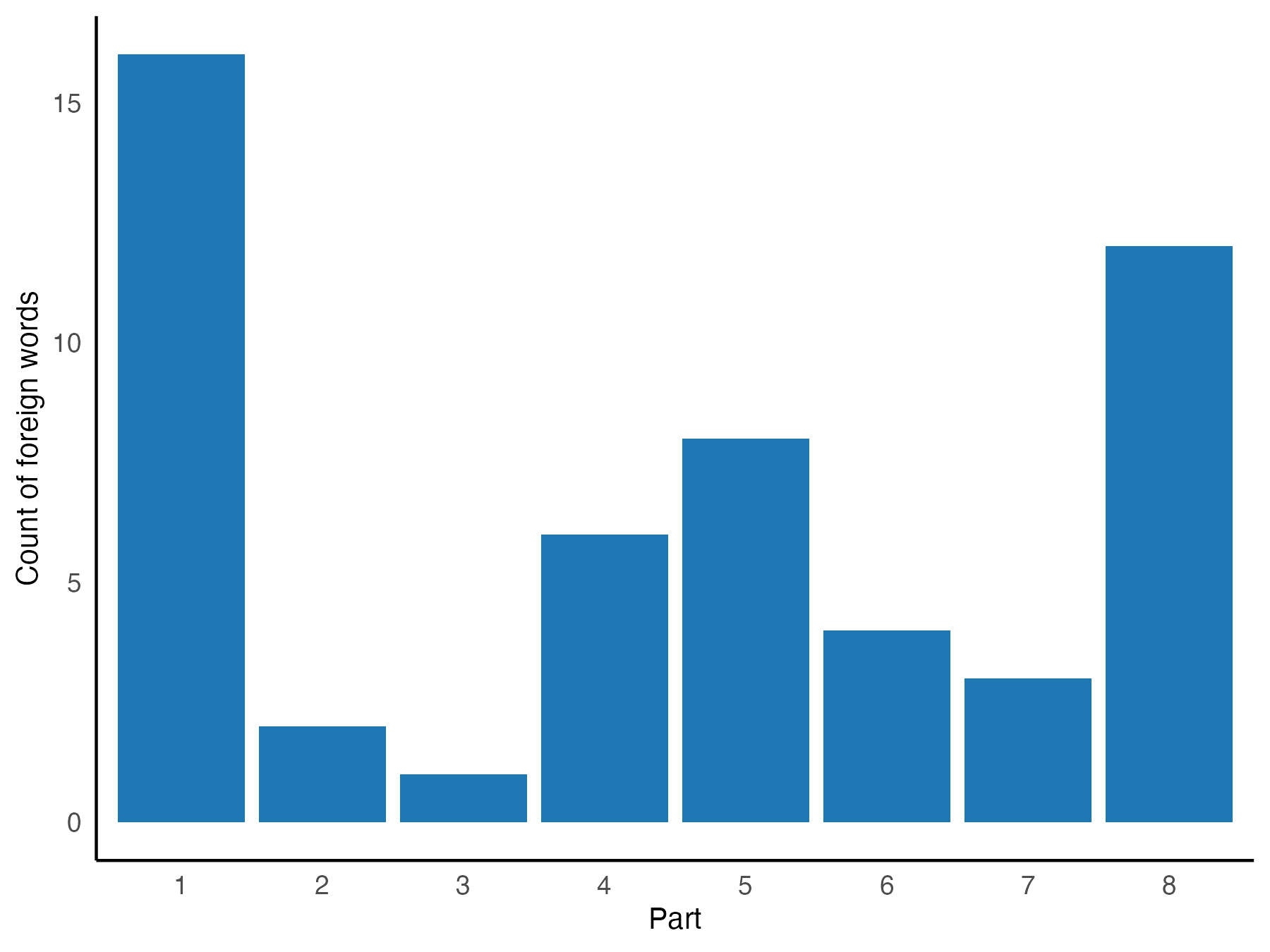}
\caption{{\bf Latin-script word usage across poem sections.}
The highest densities (words longer than 3 characters) occur in Parts~1 and~8, where the narrative unfolds in urban settings and circles of high society.}
\label{fig:fig-2}
\end{figure}

In parallel, we examined the occurrence and distribution of foreign words used by Puškin in \textit{Evgenij Onegin}. The text was scanned for non-Cyrillic tokens using a regular expression-based search. All Latin-script words of at least four characters were retained. The resulting inventory highlights both the presence and the uneven distribution of such borrowings across the poem, as shown in Fig~\ref{fig:fig-2}.

\subsection*{Tracking memory depth}
To investigate the evolution of local dependencies throughout the poem, we estimated memory depth ($\text{MD}$)~\textemdash\ as defined by the four-state Markov model~\textemdash\ across consecutive 10{,}000-character blocks. This segmentation, applied independently to the Russian and Italian V/C sequences, preserves the symbolic structure of the text while enabling blockwise comparison. The Russian text contains 107{,}168 characters, yielding ten full-length blocks plus a final partial block (7{,}000 characters) reaching Part 8, stanza 51. Given the robustness of blockwise trends to block length choice, as verified through a sensitivity analysis using shorter blocks (see S2 Appendix), the inclusion of the partial block does not materially affect slope estimation. The Italian version includes twelve full blocks (123{,}327 characters), extending to Part 8, stanza 41. Although the segmentation is not perfectly aligned, it covers nearly the entire poem in both languages and supports consistent downstream analyses.

\begin{figure}[!h]
\centering
\includegraphics[width=0.9\linewidth]{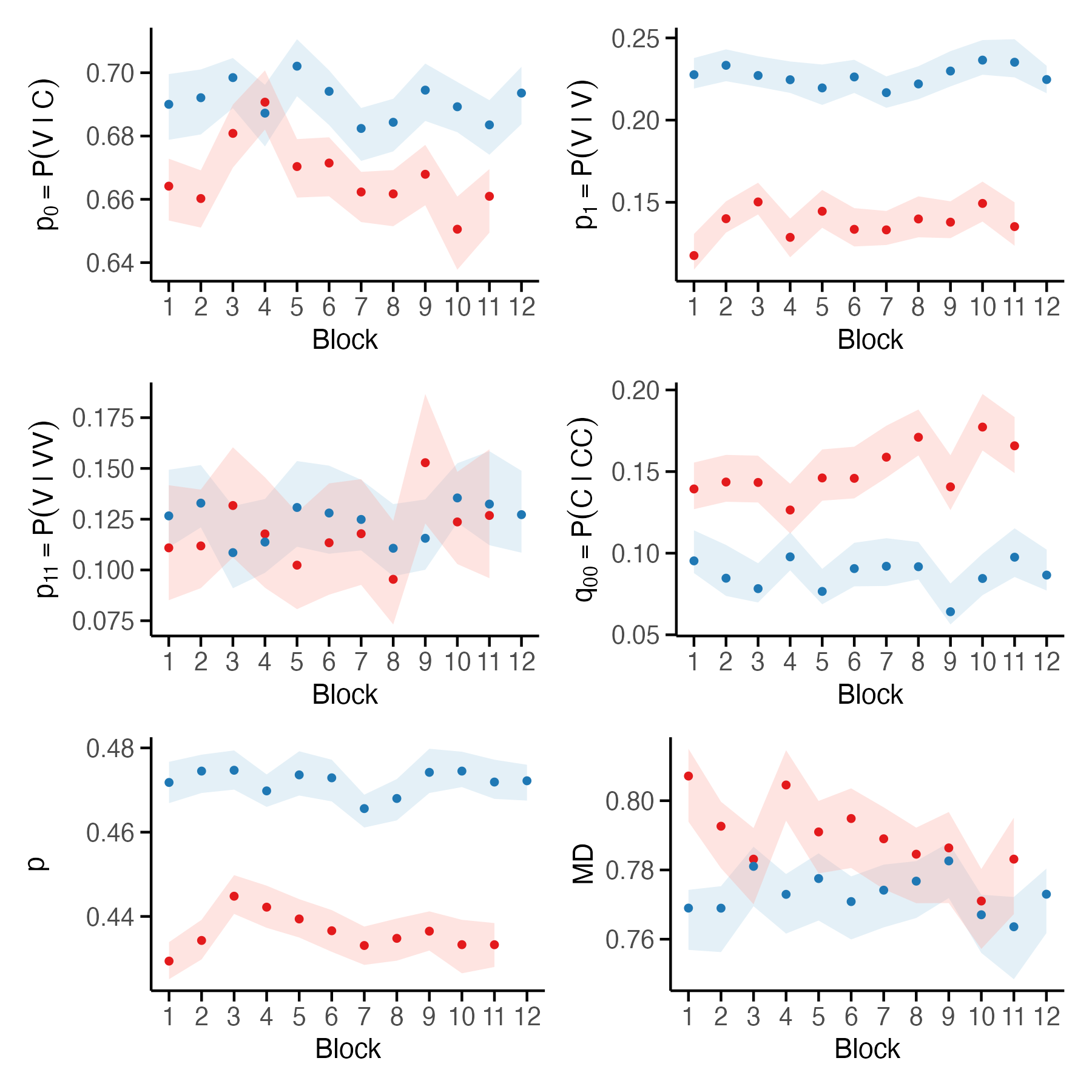}
\caption{{\bf Blockwise estimates of vowel-consonant transition probabilities and memory depth ($\text{MD}$).}
95\% bootstrap confidence intervals are shown with shaded areas. Russian (red) and Italian (blue) V/C sequences are compared across 10{,}000-character blocks.}
\label{fig:fig-3}
\end{figure}

Within each block, we computed the unigram, bigram, and trigram frequencies required to estimate $\text{MD}$, as specified in Equations~(\ref{eq:CF_simple})--(\ref{eq:MD}). Fig~\ref{fig:fig-3} displays the blockwise $\text{MD}$ estimates, along with 95\% bootstrap confidence intervals obtained via the MBB procedure.

We further assessed the association between $\text{MD}$ and its components~\textemdash\ namely, the stationary vowel probability $p$, the first-order transition probabilities $p_0$ and $p_1$, and the second-order probabilities $q_{00}$ and $p_{11}$~\textemdash\ using partial Spearman correlations. These were computed separately for Russian and Italian and are summarized in Table~\ref{tab:tab-tab2}.

\begin{table}[ht]
\centering
\footnotesize
\renewcommand{\arraystretch}{1.1}
\caption{\textbf{Partial Spearman correlations between memory depth $\text{MD}$ and model parameters.}
Estimates are based on 10{,}000-character blocks. Significance levels: $^{\ast} p < 0.05$, $^{\ast\ast} p < 0.01$, $^{\ast\ast\ast} p < 0.001$.}
\smallskip
\label{tab:tab-tab2}
\begin{tabular}{l ll ll}
\toprule
\textbf{Variable} & \multicolumn{2}{l}{\textbf{Russian}} & \multicolumn{2}{l}{\textbf{Italian}} \\
                  & $r$ & $p$ & $r$ & $p$ \\
\midrule
$p$      & $+0.13$ & $0.781$             & $-0.63$ & $0.091$ \\
$p_0$    & $-0.14$ & $0.758$             & $+0.40$ & $0.327$ \\
$p_1$    & $-0.61$ & $0.142$             & $-0.55$ & $0.160$ \\
$q_{00}$ & $-0.77$ & $0.042^{\ast}$      & $-0.93$ & $0.0007^{\ast\ast\ast}$ \\
$p_{11}$ & $-0.84$ & $0.019^{\ast\ast}$  & $-0.96$ & $0.0002^{\ast\ast\ast}$ \\
\bottomrule
\end{tabular}
\end{table}

\subsubsection*{Serial dependence and model adequacy}
The empirical autocorrelation function (ACF) was computed for each 10{,}000-character block in both the Russian and Italian V/C sequences, followed by Ljung-Box (LB) tests up to lag 10. In both cases, the LB test yielded very small $p$-values ($p < 0.0001$), allowing confident rejection of the null hypothesis of no serial dependence. These findings indicate that vowel-consonant alternation in both languages departs significantly from i.i.d. behavior, exhibiting structured temporal dependencies that support the use of low-order Markov chains as principled models for binary symbolic sequences.

\begin{figure}[!h]
\centering
\includegraphics[width=0.9\linewidth]{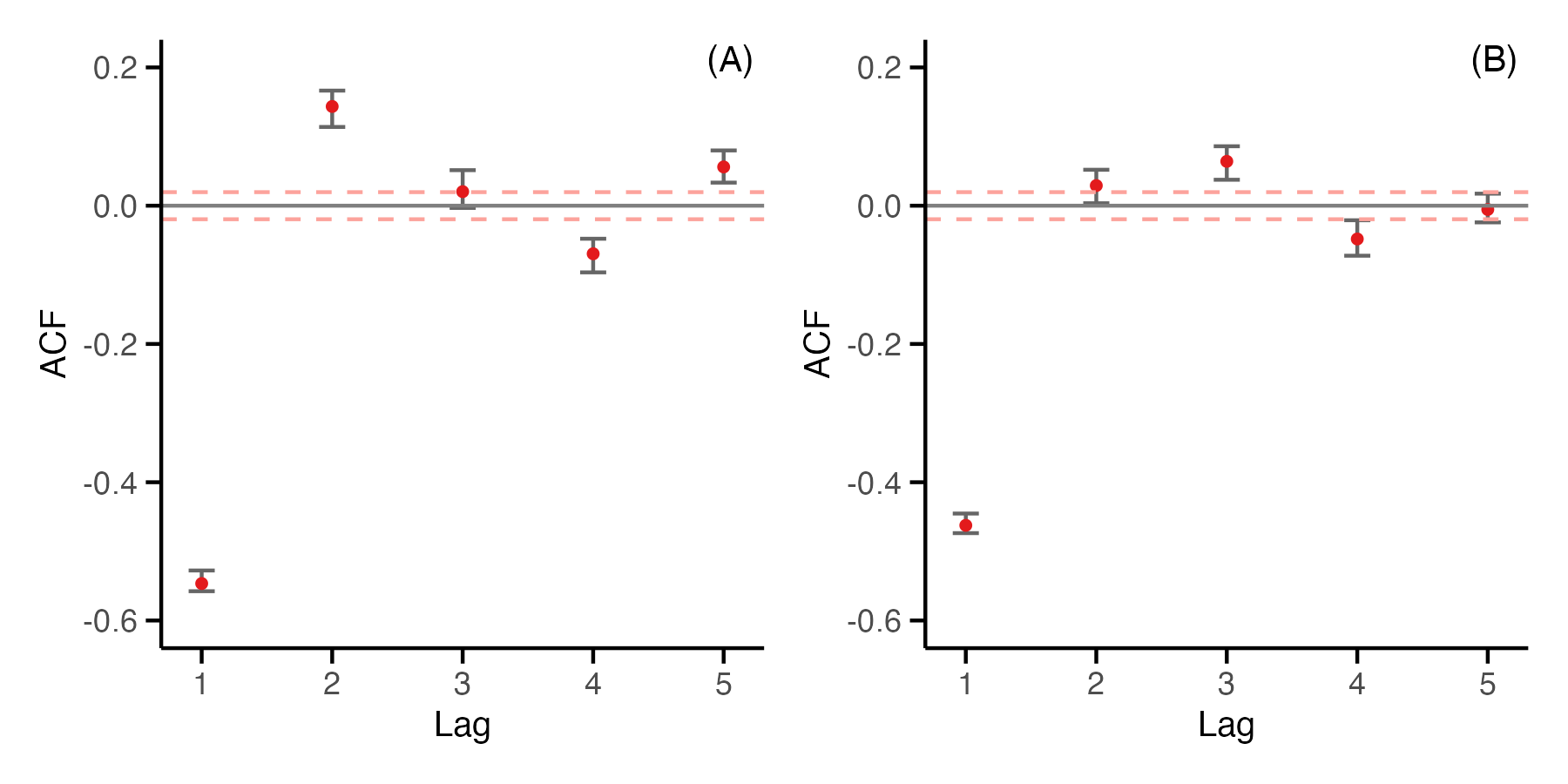}
\caption{{\bf Empirical ACF for the first 10{,}000-character block.}
Error bars denote 95\% bootstrap confidence intervals from MBB. Dashed red lines show white-noise thresholds ($\pm 1.96/\sqrt{n}$, with $n = 10{,}000$). Panel (A): Russian; Panel (B): Italian.}
\label{fig:fig-4}
\end{figure}

MBB replicates were used to estimate 95\% block-specific confidence intervals for the empirical ACF up to lag 5, as shown in Fig~\ref{fig:fig-4}.

\begin{figure}[!h]
\centering
\includegraphics[width=0.8\linewidth]{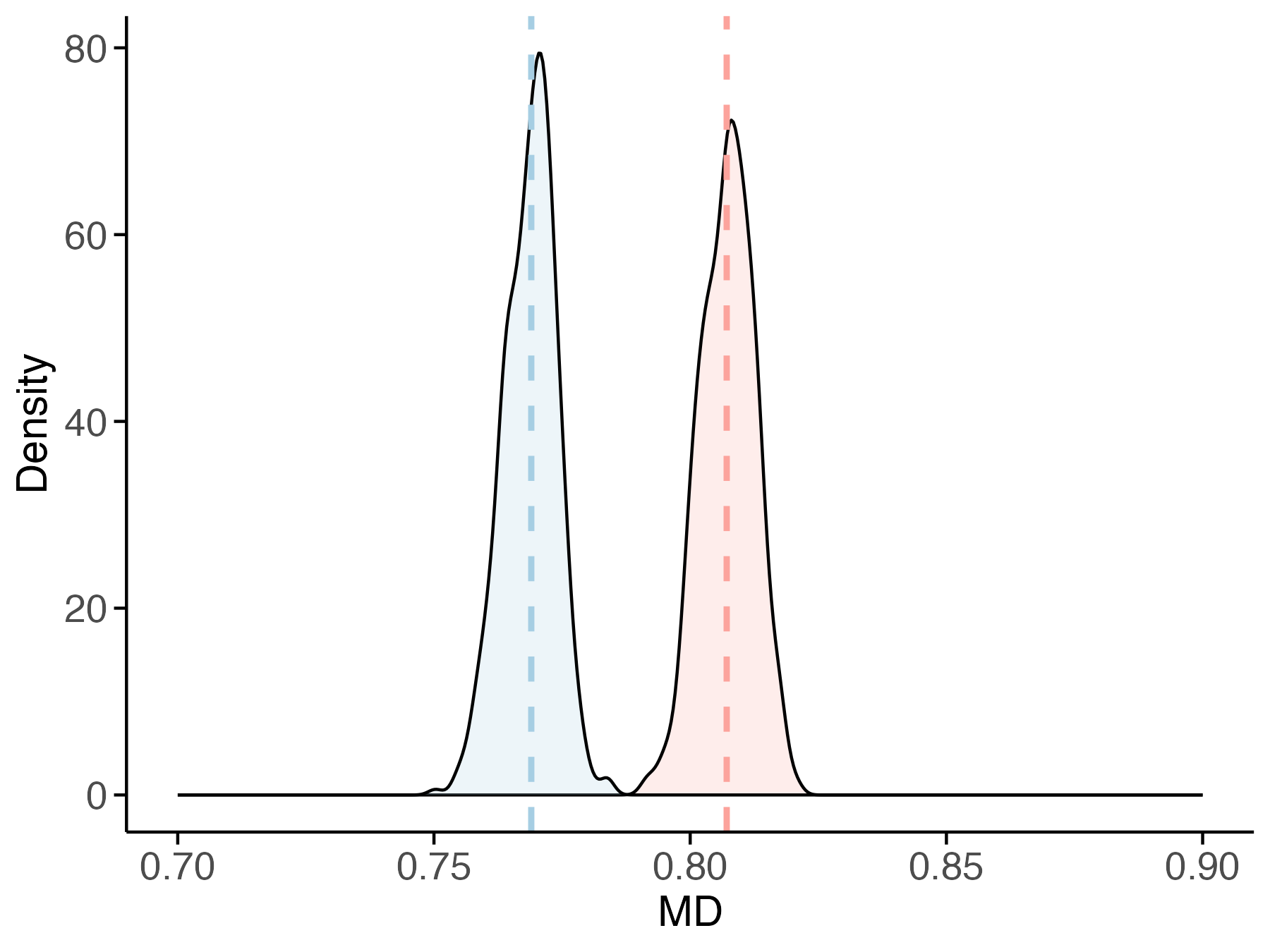}
\caption{{\bf Bootstrap distributions of $\text{MD}$.}
The four-state Markov model is fitted to the Russian and Italian V/C sequences. The dashed line marks the empirical value from the original text. Russian (red); Italian (blue).}
\label{fig:fig-5}
\end{figure}

To evaluate the adequacy of the four-state Markov chain as a generative model, we simulated 500 sequences of 10{,}000 characters using transition probabilities estimated from the first empirical block. For both languages, the distribution of simulated $\text{MD}$ values closely matches the empirical benchmark. As shown in Fig~\ref{fig:fig-5}, the empirical $\text{MD}$ (dashed line) falls well within the 95\% range of simulated values. Median discrepancy values were small for both corpora (Russian: 0.0058, 95\% bootstrap interval [0.0042, 0.0078]; Italian: 0.0035 [0.0024, 0.0052]), indicating close agreement between empirical and simulated trigram distributions and supporting the adequacy of the four-state model as a generative representation of local vowel-consonant structure.

\subsubsection*{Evolution of memory depth over text}
To assess whether $\text{MD}$ evolves differently across the Russian original and its Italian translation, we fitted a linear model with interaction: 
$\texttt{MD} \sim \texttt{block} \times \texttt{source}$. 
This formulation allows testing for differences in trend between the two languages, with the following components: 
(a) an intercept (baseline $\texttt{MD}$ for \texttt{Italian}), 
(b) a main effect of block position ($\texttt{block}$, trend in \texttt{Italian}), 
(c) a main effect of language ($\texttt{source}$), and 
(d) an interaction term ($\texttt{block} \times \texttt{Russian}$), which captures deviations of the Russian slope from the Italian one.

To assess the significance and stability of the interaction term, we applied a nonparametric bootstrap procedure: the model was re-fitted on each of 1{,}000 MBB replicates, and the full distribution of estimated coefficients was extracted. This approach avoids strong parametric assumptions, particularly relevant given the limited number of text blocks and the intrinsic temporal dependence of the sequences. Instead of relying on asymptotic standard errors, confidence intervals were derived from the empirical bootstrap distributions.

\begin{figure}[!h]
\centering
\includegraphics[width=0.9\linewidth]{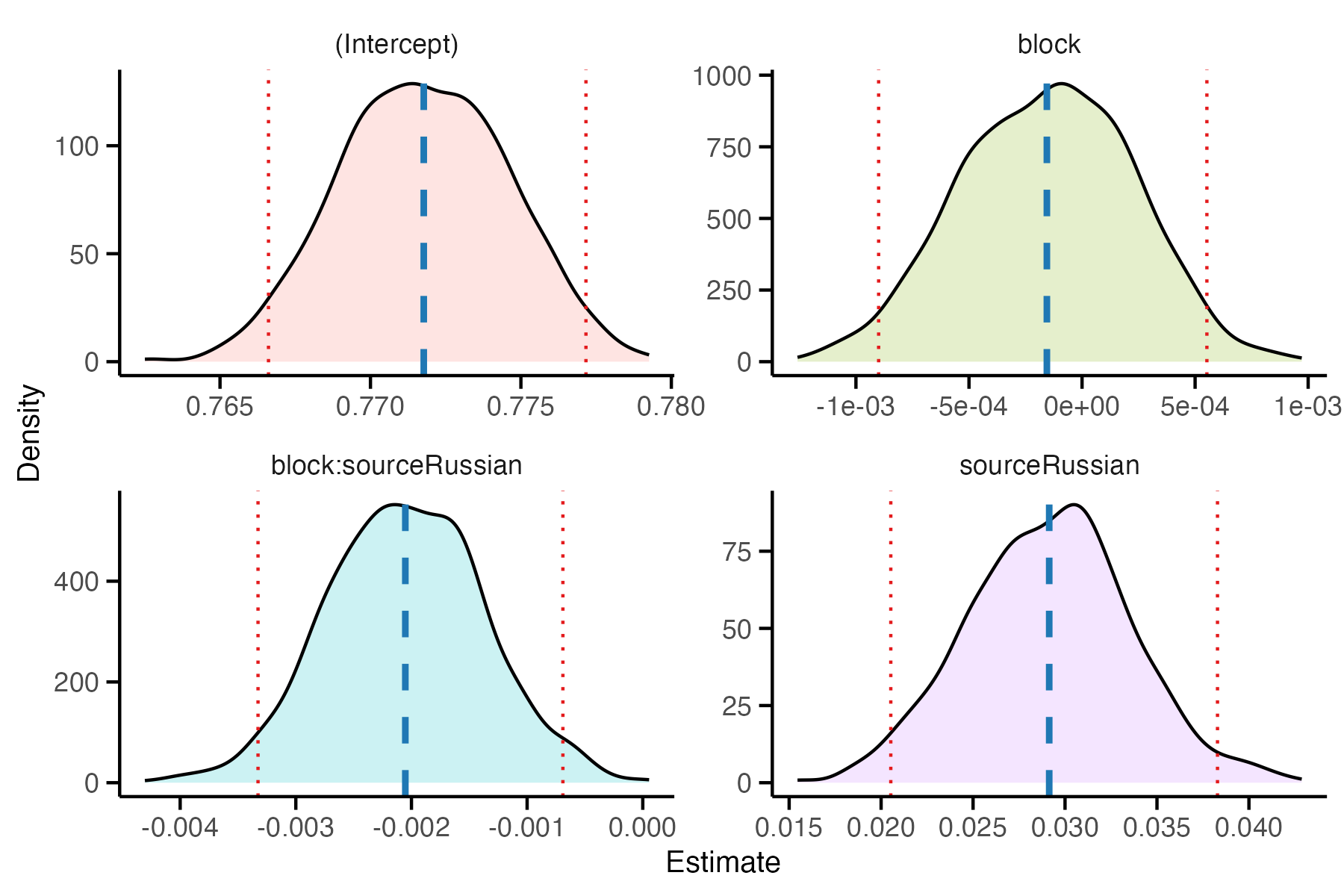}
\caption{{\bf Empirical bootstrap distributions of the linear model coefficients.}
Dashed blue lines denote the mean estimate; dotted red lines mark the 95\% confidence interval. The interaction term is strictly negative, indicating diverging $\text{MD}$ dynamics.}
\label{fig:fig-6}
\end{figure}

As shown in Fig~\ref{fig:fig-6}, all model coefficients were stable across replicates. In particular, the interaction term is consistently negative, with a 95\% confidence interval that excludes zero, thereby supporting the hypothesis of diverging memory dynamics.

\begin{figure}[!h]
\centering
\includegraphics[width=0.8\linewidth]{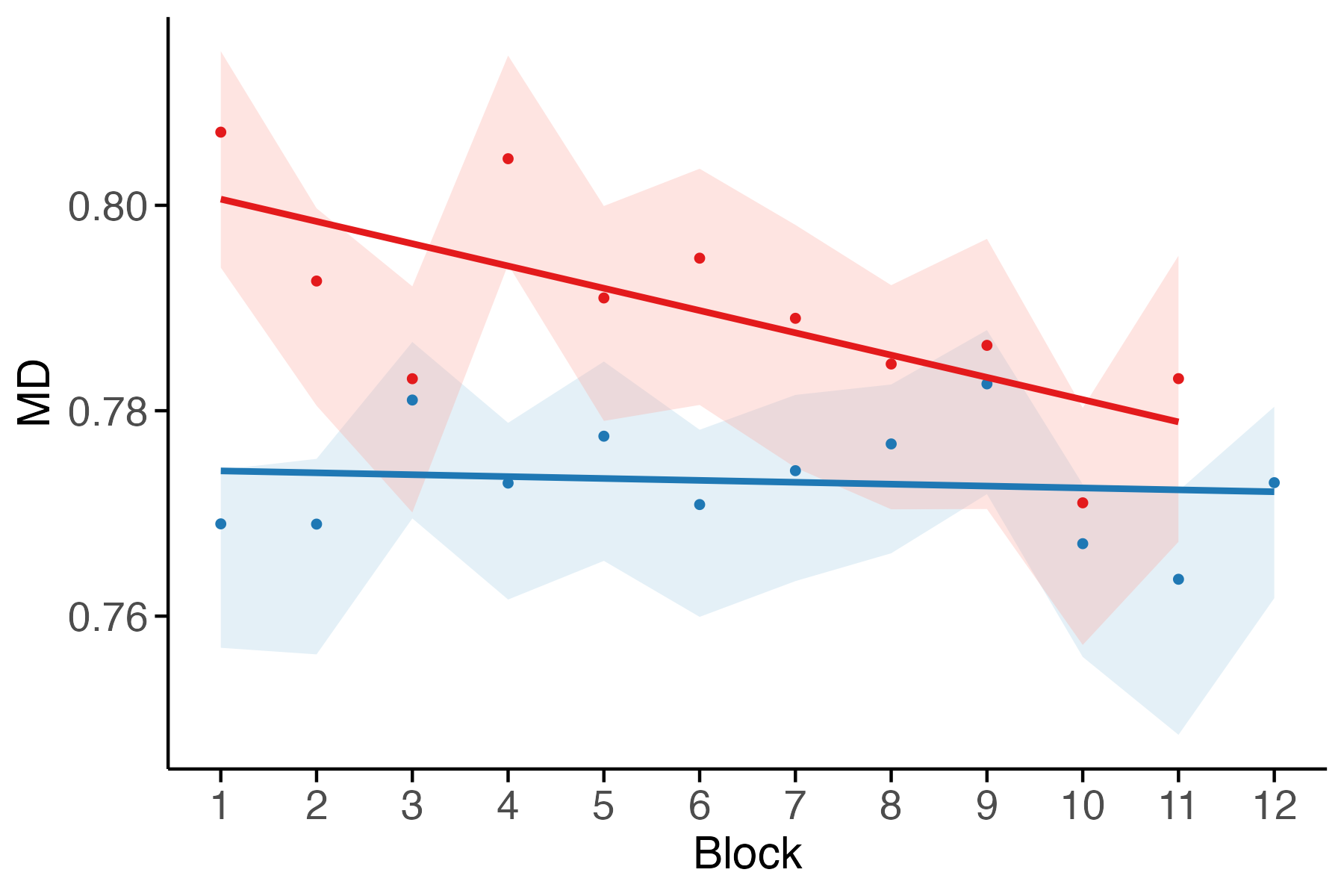}
\caption{{\bf Block-wise estimates of $\text{MD}$.}
95\% bootstrap confidence intervals are shown with shaded areas. The regression lines reveal a decreasing trend in the Russian text (red), in contrast to the stable trajectory observed in the Italian translation (blue).}
\label{fig:fig-7}
\end{figure}

Fig~\ref{fig:fig-7} illustrates the fitted trajectories across blocks: while $\text{MD}$ remains stable in the Italian translation (blue), the Russian original (red) exhibits a clear decreasing trend. Surrogate-data tests confirmed that this divergence is not an artifact of local character clustering or marginal frequencies: reshuffled versions of the Russian sequence failed to reproduce the observed monotonic trend. Moreover, repeating the same regression using surrogates in place of the Italian sequence yielded virtually identical interaction estimates and confidence intervals, further reinforcing the robustness of the observed effect.

\subsection*{Probing the phonological structure}
Having identified second-order transition probabilities as key drivers of vowel-consonant alternation, we turned to a finer-grained analysis of the trigram structures underlying
this effect. We focus in particular on the two configurations most strongly associated with $\text{MD}$ (VVV and CCC) as revealed by partial Spearman correlation analyses across both corpora. Their complementary counterparts (VVC and CCV) exhibit analogous associations of opposite sign. Since the contrast lies in the degree of continuity of vocalic or consonantal patterns, we refer to VVV and CCC as \textit{persistent trigrams}, and to VVC and CCV as \textit{alternating trigrams}. Table~\ref{tab:tab-tab3} reports the raw counts of the four vowel-consonant trigram classes in the Russian and Italian corpora. While alternating patterns dominate in both languages, persistent trigrams (VVV and CCC) remain sufficiently frequent to support trend estimation and contextual analysis.

\begin{table}[ht]
\centering
\caption{Raw counts of vowel--consonant trigram classes in the Russian and Italian corpora.}
\smallskip
\label{tab:tab-tab3}
\smallskip
\begin{tabular}{lrr}
\toprule
Trigram class & Russian & Italian \\
\midrule
CCC & 3029 & 1744 \\
CCV & 17058 & 18345 \\
VVC & 5660 & 11605 \\
VVV & 762 & 1656 \\
\bottomrule
\end{tabular}
\end{table}

\subsubsection*{Investigating high-impact trigrams}
To examine how these phonotactic configurations are lexically realized, we scanned the V/C-encoded sequences for all instances of persistent and alternating trigrams in both corpora. Spaces were ignored during matching, but original character positions were preserved. For each match, we extracted its lexical context~\textemdash\ bounded by the nearest spaces~\textemdash\ and recorded its location within the poem (part, stanza, and line).

Characters excluded from the V/C encoding, such as apostrophes, the Russian hard sign (ъ), and soft sign (ь) were retained in the reconstructed context to preserve phonological and morphological integrity. Each instance was then classified by trigram type (persistent or alternating) and by whether it occurred in either single-word or multi-word context.

To move beyond surface-level phonotactic patterns, we defined a set of phonological probes based on three post hoc criteria:

1. Statistical association with text progression: measured by fitting a linear model to the blockwise frequency of each trigram and computing the Spearman $p$-value;

2. Morphological simplicity: assessed via identification of single-word occurrences that allow robust lemmatization and syntactic analysis;

3. Semantic coherence: established through manual inspection of lemma consistency and thematic recurrence across matched contexts.

Most candidate probes shown in Table~\ref{tab:tab-tab4} did not meet all three criteria. Among the Italian candidates, "lch" occurred only in the single-word occurrence \textit{qualche} (Italian for \textit{some} or \textit{a few}); "ioe" and "eal" were rare and semantically heterogeneous; and while "nte" appeared frequently (mostly in adverbs and adjectives), it lacked clear thematic cohesion. A similar situation applied to the Russian candidates "ыйв", "ыйп", and "тпр", which either lacked single-word occurrences or occurred only sporadically. The trigram "ает" was frequent but confined to verbal endings, limiting its semantic specificity.

\begin{table}[ht]
\centering
\footnotesize
\renewcommand{\arraystretch}{1.1}
\caption{\textbf{Candidate phonological probes}.
Trigrams exhibiting significant correlations with block position are shown. The column \textbf{Match} indicates whether the trend aligns with expectations from the $\text{MD}$ dynamics.}
\smallskip
\label{tab:tab-tab4}
\begin{tabular}{@{}lllll@{}}
\toprule
\textbf{Language} & \textbf{Probe} & \textbf{Direction} & \textbf{Pattern} & \textbf{Match} \\
\midrule
\textbf{Russian}  & "ает" & increasing  & alternating & no  \\
                  & "вст" & increasing  & persistent  & yes \\
                  & "тра" & decreasing  & alternating & yes \\
                  & "тпр" & increasing  & persistent  & yes \\
                  & "ыйв" & decreasing  & alternating & no  \\
                  & "ыйп" & increasing  & alternating & no  \\
\textbf{Italian}  & "eal" & decreasing  & alternating & yes \\
                  & "ioe" & decreasing  & persistent  & no  \\
                  & "lch" & decreasing  & persistent  & no  \\
                  & "nte" & decreasing  & alternating & yes \\
\bottomrule
\end{tabular}
\end{table}

In summary, only two probes, namely "вст" and "тра", satisfied all criteria and were retained for further analysis. As shown in the subsequent section, only “вст” ultimately proved suitable for detailed lexical and narrative investigation, whereas “тра” was excluded after closer examination.

\subsubsection*{Phonological probes and associated lemmas: the case "вст"}
To explore how phonological patterns may map onto recurring narrative or emotional motifs, we conducted a focused case study on the consonantal trigram "вст", selected from Table~\ref{tab:tab-tab4} based on its strong upward trend and classification as a persistent pattern. Among the top trigram patterns in the Russian corpus, forms such as "ста" (7.2\%, rank 1), "при" (6.4\%, rank 2), and "тра" (3.6\%, rank 14) followed a steep Zipfian drop \cite{piantadosi2014}, reflecting broad syntactic and narrative usage. Just below this threshold lies "вст" (2.3\%, rank 19)~\textemdash\ frequent enough to support statistical analysis, yet rare enough to retain discriminative potential.

Initial tracking of all "вст" occurrences revealed a heterogeneous set of forms. To enhance interpretability, we restricted analysis to matches of single-word occurrences, which offered cleaner lemmatization and syntactic consistency. Each match was processed via UDPipe and manually reviewed to correct tagging errors (see S3 Appendix, Table A1).

To identify semantically cohesive clusters, we applied the SnowballC stemming algorithm to the corrected lemmas. Although not used directly for tagging or classification, the resulting stem clusters helped validate the thematic consistency of lexical groupings (see S3 Appendix, Table A2). These were classified into two dominant categories: \textit{encounter} (e.g., встр-, вступ-, здравст-) and \textit{emotion} (e.g., чувств-, предчувств-, девств-, бесчувст-). Thematic labels were then assigned accordingly, enabling recomputation of blockwise frequencies by semantic group.

The data shown in Table~\ref{tab:tab-tab5} confirm a significant upward trend in frequency for encounter-related forms, in parallel with the decrease in $\text{MD}$ across the poem. Emotion-related forms show a weaker, non-significant trend, though they remain thematically relevant. Borderline cases such as вздохов страстных (\textit{passionate sighs}) or порыв страстей (\textit{surge of passion}) often involve adjacent tokens within a coherent emotional frame, reinforcing the overall signal.

\begin{table}[ht]
\centering
\footnotesize
\renewcommand{\arraystretch}{1.1}
\caption{\textbf{Spearman correlation between block index and frequency of the phonological probe.} Total number of matches ($n$) and $p$-values for the phonological probe "вст", by semantic category.}
\smallskip
\label{tab:tab-tab5}
\begin{tabular}{lll}
\toprule
\textbf{category} & \textbf{occurrences ($n$)} & \textbf{$p$-value} \\
\midrule
emotion                & 55   & 0.174 \\
encounter              & 29   & 0.00021$^{\ast\ast\ast}$ \\
emotion + encounter    & 84   & 0.00418$^{\ast\ast}$ \\
all probe matches      & 111  & 0.00193$^{\ast\ast}$ \\
\bottomrule
\end{tabular}
\end{table}

Further analysis focused on the thematic anchoring of the root встр, central to several verbs of encounter (e.g., встретить, встречать). We examined its co-occurrence with the protagonists’ names, Tat’jana Larina and Evgenij Onegin, mentioned a total of 269 times in the poem. In 25\% of these instances, at least one name appears in the same stanza as the probe. Conversely, thematic forms linked to встр occur in 30\% of the stanzas containing “вст”. Restricting attention to these sections, we observed a modest yet statistically significant correlation between name mentions and thematic presence (Spearman’s $\rho = 0.23$, $p < 0.05$), consistent with a non-random association between them.

By contrast, the alternating trigram "тра", though initially promising, failed to yield stable results. Lexical items such as страсть (\textit{passion}), страх (\textit{fear}), and страшный (\textit{terrible}) were too sparse or contextually dispersed to support robust trend analysis. Moreover, its high raw frequency~\textemdash\ reflected in an elevated Zipfian rank~\textemdash\ diluted its contrastive power. Unlike "вст", whose discriminative signal arises from patterned and thematically coherent recurrences, "тра" appears ubiquitous but semantically diffuse.

\section*{Discussion}
The marked Gaussianity observed in line-level character and word counts can be traced back to the metrical constraints embedded in the poem’s verse structure. As discussed in the Introduction, Puškin’s tetrametrical iambs and Ghini’s unrhymed nine-syllable lines impose strict formal templates that shape rhythm and, indirectly, the distribution of graphemes and lexical items. In this context, the tight clustering of word and character counts arises not from statistical averaging, but from the deliberate stylization of the poetic line. The comparatively lower average counts in Russian further reflect the typological compactness of the language, whose rich inflection and flexible word order contrast with the more analytic structure of Italian, where grammatical relations are often conveyed through separate function words and fixed syntactic patterns \cite{pushkin2021}.

Puškin’s style is famously heterogeneous, and his deliberate use of code-switching (especially into French, but also German, Italian, or English) serves multiple rhetorical and narrative functions, from irony to social satire. Although rare in absolute terms (< 0.25\% of total words), such borrowings tend to cluster in Parts 1 and 8~\textemdash\ the poem’s urban frame~\textemdash\ highlighting the social and stylistic codes of high society (Fig~\ref{fig:fig-2}). While these insertions are not expected to significantly affect the Markovian memory of the text, preserving multilingual characters is considered essential to retain the stylistic heterogeneity of the original during binary vowel-consonant encoding. The same principle is applied to the Italian version, where most foreign expressions are preserved or carefully adapted, often maintaining both their stylistic role and foreign flavor. In addition, Russian words, especially proper nouns, are frequently transliterated with diacritics, in order to preserve their structure across writing systems.

\subsection*{Tracking memory depth}
Blockwise modeling of vowel-consonant alternation reveals a clear divergence between the Russian and Italian versions of \textit{Evgenij Onegin} (Fig \ref{fig:fig-3}). In the Russian corpus, $\text{MD}$~\textemdash\ a second-order Markovian measure of local V/C cohesion~\textemdash\ shows a significant downward trend across blocks, suggesting a gradual loosening of sequential constraints as the poem unfolds. By contrast, the Italian translation displayed stable $\text{MD}$ values with no discernible trend. As shown by the partial correlation results in Table 2, second-order transition probabilities~\textemdash\ most notably the probability of a vowel following two vowels and that of a consonant following two consonants~\textemdash\ are strongly associated with $\text{MD}$, in line with its theoretical formulation of Equations~(\ref{eq:CF_simple})--(\ref{eq:MD}).

Bootstrap-based resampling confirms the robustness of $\text{MD}$ estimates, which remained tightly centered around their empirical values. This supports the interpretation that the observed trajectory reflects genuine stylistic development rather than artifacts of sampling variability. Furthermore, local peaks and troughs in $\text{MD}$ frequently align with key narrative transitions, suggesting that stylometric fluctuations may accompany~\textemdash\ and possibly reflect~\textemdash\ the unfolding of the plot.

As shown in Figures~\ref{fig:fig-4}--\ref{fig:fig-5}, validation of the Markovian model through autocorrelation analysis and synthetic sequence generation further confirms that the four-state structure is adequate to reproduce both local transition patterns and the global dispersion of vowel-consonant alternation observed in the original texts. Initially introduced as a compact descriptive proxy, the four-state chain can be regarded as a valid generative model that is capable of capturing the structural regularities observed in the data and reproducing them through simulation. These findings validate its use as a baseline for statistical resampling and inference.

The most salient result of our block-level analysis is the divergence in $\text{MD}$ values between the two texts. Bootstrap modeling confirms that this divergence is not an artifact of sampling variability: the interaction term in the regression model remains consistently negative across resampled datasets, and its confidence interval excludes zero (Fig~\ref{fig:fig-6}). The Russian original exhibits a statistically robust decreasing trend, while the Italian translation remains flat (Fig~\ref{fig:fig-7}).

\subsection*{Probing the phonological structure}
To probe the sequential dynamics of the symbolic sequences, we analyze second-order trigram patterns (VVV, VVC, CCV, CCC) along with the associated transition probabilities~\textemdash\ specifically, $p_{11} = P(V \mid VV)$ and $q_{00} = P(C \mid CC)$~\textemdash\ as well as their complementary forms. In the Russian corpus, the frequency of persistent trigrams (VVV and CCC) increases across blocks, while that of alternating trigrams (VVC and CCV) decreases. This pattern suggests growing vocalic and consonantal clustering, enhancing short-range predictability and contributing to the observed decline in $\text{MD}$. These trends are confirmed by bootstrap estimates, which tracked closely with the blockwise evolution of $\text{MD}$.

Rather than relying on fully automated approaches, we adopt a hybrid strategy combining statistical filtering, exploratory visualization, and manual inspection. This proves well-suited to a corpus of moderate size, where dense phonotactic clusters are rare and their emergence often narratively charged. Among the patterns identified, the trigram "вст" stands out for its significant upward trend and thematic coherence, particularly in contexts limited to single-word occurrences, where it consistently marked moments of narrative tension and interpersonal encounter.

In the Russian text of \textit{Evgenij Onegin}, "вст" does not emerge as a dominant lexical item, but as a persistent phonological pattern whose distribution is structured rather than random. While its overall frequency remains moderate, its recurrent occurrence is statistically aligned with sections of the poem involving Tat’jana and Evgenij, suggesting a non-random association between this phonological pattern and key narrative contexts.

\subsection*{Reflections on method and scope}
All inferences drawn in this study should be understood as exploratory. Correlation does not imply causation, and the associations observed, whether supporting phonological probes or block-level trends, are best interpreted as hypothesis-generating signals, inviting further linguistic and narrative inquiry.

The adopted pipeline combined symbolic encoding, fixed-block modeling (10{,}000-character windows), and lexically curated probes; it was intentionally simple and reproducible. Unlike modern NLP approaches that rely on expressive representations~\textemdash\ such as bag-of-words, multi-hot, or contextual embeddings~\textemdash\ our binary vowel-consonant model operates strictly at the graphemic surface. It discards lexical identity and syntax, reducing the text to a sparse rhythmic sequence. As such, it cannot access latent semantic structures or uncover topic-level organization, as more complex vector-based encodings might allow.

Yet this minimalism allows certain signals to emerge despite the coarseness of the encoding. The phonological probe that surfaced~\textemdash\ namely "вст"~\textemdash\ emerged as a recurrent graphemic marker. That such cues were detectable at all may be due to two converging factors: a language (Russian) that is structurally amenable to Markovian modeling (e.g., due to phonotactic regularity and morphological density) \cite{petruszewycz1983}, and a narrative structure in which the recurrence of specific morphophonological patterns is observable across the text.

\section*{Conclusion}

This study shows that even minimalist modeling, when grounded in linguistic and literary structure, can uncover statistically robust and thematically meaningful patterns. A simple four-state Markov model applied to vowel-consonant sequences extracted from the Russian and Italian versions of \textit{Evgenij Onegin}, combined with block-wise analysis and bootstrap validation, successfully captured a progressive shortening of phonotactic memory in the Russian corpus. This result was corroborated by second-order transition probabilities and the emergence of targeted phonological probes.

That such patterns could be detected using a binary encoding suggests that some signals in literary texts are strong enough to surface even through deliberately reduced lenses. 
This outcome is consistent with the view that Markov chains~\textemdash\ despite their simplicity~\textemdash\ might still hold untapped potential for literary analysis \cite{petruszewycz1983}. These findings could be confirmed and even extended by adopting richer encodings~\textemdash\ such as bag-of-words representations coupled with topic modeling (e.g., Latent Dirichlet Allocation)~\textemdash\ to uncover latent semantic structures and longer-range narrative cues. This remains a promising direction for future inquiry and forms part of our ongoing research agenda.

\bibliographystyle{plos2015}

\end{document}